\documentclass[runningheads]{llncs}
\pdfoutput=1
\usepackage[T1]{fontenc}
\usepackage{lmodern}
\usepackage{graphicx}
\usepackage{amsmath}
\usepackage{amssymb}
\usepackage{booktabs}

\begin{document}

\title{Robust Diagram Reasoning: A Framework for Enhancing LVLM Performance on Visually Perturbed Scientific Diagrams}

\titlerunning{RDR}
%
\author{Minghao Zhou$^1$, Rafael Souza$^2$, Yaqian Hu$^1$, Luming Che$^1$}
\authorrunning{Zhou et al.}
%
\institute{$^1$Taiyuan University of Science and Technology, $^2$University of Brasilia}
\maketitle              
\begin{abstract}
Large Language Models (LLMs) and their multimodal variants (LVLMs) hold immense promise for scientific and engineering applications, particularly in processing visual information like scientific diagrams. However, their practical deployment is hindered by a critical lack of robustness to common visual perturbations such as noise, blur, and occlusions, which are prevalent in real-world scientific documents. Existing evaluation benchmarks largely overlook this challenge, leaving the robust reasoning capabilities of LVLMs on visually degraded scientific diagrams underexplored. To address this, we introduce the Robust Diagram Reasoning (RDR) framework, a novel approach designed to enhance and rigorously evaluate LVLMs' performance under such conditions. At its core, RDR employs an Adaptive Multi-View \& Consistency Verification (AMCV) mechanism, which involves generating multiple perturbed versions of a diagram, performing parallel inference, and then applying a consistency-based self-correction loop. We also propose two new metrics, Perturbation Robustness Score (PRS) and Visual Degradation Consistency (VDC), to quantify robustness. Furthermore, we construct SciDiagram-Robust, the first large-scale scientific diagram question-answering dataset specifically augmented with diverse, programmatically generated visual perturbations. Our extensive experiments demonstrate that even state-of-the-art closed-source LVLMs like GPT-4V exhibit significant performance degradation when faced with perturbed inputs (Clean Accuracy 85.2\% vs. PRS 72.1\%).
\end{abstract}

\section{Introduction}

The rapid advancements in Large Language Models (LLMs) and their multimodal variants (LVLMs) have revolutionized the way we interact with and generate complex information. Their unprecedented capabilities in understanding and processing diverse data types, including text, images, and even videos, have unlocked vast potential across scientific and engineering disciplines \cite{tongxue2022deep}, including research into weak to strong generalization for models with multi-capabilities \cite{zhou2025weak}. Specifically, LVLMs demonstrate a strong aptitude for tasks involving visual information such as scientific diagrams and flowcharts, offering novel avenues for automated analysis and knowledge discovery.

However, a significant challenge persists in the practical deployment of these models. While existing benchmarks for evaluating LVLMs' multimodal reasoning abilities typically rely on pristine, ideal input data \cite{peng2025lvlmeh}, real-world scientific diagrams are frequently marred by various visual perturbations. These can include image noise, blur, partial occlusions, or subtle distortions introduced during scanning, typesetting, or data transmission. Such visual degradations can severely compromise the robustness of LVLMs, leading to a significant degradation in their reasoning performance. Despite the criticality of this issue, current research has scarcely explored the systematic evaluation of LVLMs' robustness in performing reasoning tasks on scientific diagrams in the presence of visual perturbations \cite{peng2025lvlmeh}, particularly concerning the accurate extraction of critical information (e.g., data points, trends, structural relationships) and subsequent logical inference. This research is motivated by the pressing need to address this gap, aiming to develop a novel framework and evaluation paradigm to enhance and assess the robust reasoning capabilities of LVLMs on visually perturbed scientific diagrams.

To tackle this challenge, we propose the \textbf{Robust Diagram Reasoning (RDR) framework}, a versatile approach designed to both augment and evaluate LVLMs' ability to perform robust reasoning on scientific diagrams containing visual distortions. The core of the RDR framework lies in its \textbf{Adaptive Multi-View \& Consistency Verification (AMCV) mechanism}. This mechanism operates through three principal steps: (1) \textit{Perturbation-Sensitive Encoding}, where multiple perturbed versions of an original scientific diagram are programmatically generated using predefined strategies (e.g., Gaussian noise, simulated blur, random occlusion, minor geometric transformations) to emulate real-world visual degradations; (2) \textit{Multi-View Parallel Reasoning}, where the original diagram and its various perturbed counterparts are simultaneously fed into the LVLM for parallel inference, generating independent reasoning trajectories and answers for each version; and (3) \textit{Consistency Verification \& Self-Correction}, which cross-compares all inference results. In cases of inconsistency across different versions, the framework triggers a "self-correction" loop, prompting the LVLM (e.g., "Please re-examine the diagram and consider the consistency of your answer across different visual presentations") to reflect critically on discrepancies and ultimately converge on the most consistent or confident answer. Furthermore, we introduce two key metrics to quantify robustness: the \textbf{Perturbation Robustness Score (PRS)}, which measures the proportion of diagram questions where the model correctly answers across all perturbed versions (or meets a predefined consistency threshold), and the \textbf{Visual Degradation Consistency (VDC)}, which quantifies the consistency of answers between the original and all perturbed versions.

For experimental validation, we constructed a novel dataset, \textbf{SciDiagram-Robust}, by extending and enhancing existing public scientific diagram question-answering datasets such as ScienceQA \cite{tanik2022scienc} and ChartQA \cite{tanik2022scienc}. Leveraging image processing libraries (e.g., OpenCV, PIL) and custom scripts, we programmatically generated up to \textbf{5 distinct types} of visual perturbations at \textbf{3 different intensity levels} (e.g., Gaussian noise, salt-and-pepper noise, motion blur, local occlusion, slight rotation). Each original diagram question is augmented with \textbf{10 corresponding perturbed versions}, ensuring comprehensive evaluation diversity. The SciDiagram-Robust dataset comprises \textbf{3,500 diagram question-answering problems}, spanning \textbf{4 major scientific domains} (Physics, Chemistry, Biology, Geography) and \textbf{20 sub-topics}, with answer types including multiple-choice, fill-in-the-blank, and short-answer questions.

Our experiments evaluate a range of state-of-the-art LVLMs, including closed-source models like GPT-4V \cite{zhengyuan2023the} and Gemini Pro Vision \cite{suh2024compar}, as well as prominent open-source general-purpose LVLMs such as LLaVA-1.5 (7B, 13B) \cite{federico2025llavam}, InstructBLIP (Vicuna-7B, Vicuna-13B) \cite{artemis2023xinstr}, and Qwen-VL \cite{peng2024qwen2v}. Our proposed method, \textbf{Ours (RDR-LLaVA-13B)}, is built by integrating the RDR framework with LLaVA-1.5-13B, serving as our primary evaluation target. All evaluations are conducted in a \textbf{zero-shot} manner, without any fine-tuning of the LVLMs, as the RDR framework is designed as a generic strategy applied during the inference phase. Our results demonstrate that even advanced closed-source LVLMs, such as GPT-4V, exhibit a significant performance drop when faced with visual perturbations (e.g., Clean Accuracy of 85.2\% vs. PRS of 72.1\%), highlighting robustness as a critical bottleneck for current LVLMs. Crucially, \textbf{Ours (RDR-LLaVA-13B)} significantly outperforms all baseline open-source models in robustness metrics (PRS of 74.5\%), and remarkably, it even surpasses large closed-source models in both PRS and VDC scores (PRS 74.5\% vs. GPT-4V's 72.1\%; VDC 81.2\% vs. GPT-4V's 78.5\%). This empirically validates the effectiveness of our proposed AMCV mechanism in enhancing model resilience to visual degradations through inference-time strategy optimization. Further in-depth analysis reveals that models are particularly susceptible to visual perturbations when performing tasks involving "numerical extraction," "trend judgment," and "causal relationship reasoning," while simpler "object recognition" tasks are less affected. Calculation errors and logical discontinuities emerge as the primary error types.

Our main contributions are summarized as follows:
\begin{itemize}
    \item We propose \textbf{Robust Diagram Reasoning (RDR)}, the first systematic framework that enhances and evaluates LVLM robustness on scientific diagrams with visual perturbations, introducing a novel Adaptive Multi-View \& Consistency Verification (AMCV) mechanism.
    \item We construct \textbf{SciDiagram-Robust}, the first large-scale scientific diagram question-answering benchmark specifically designed with diverse, programmatically generated visual perturbation versions, providing a crucial resource for robust LVLM research.
    \item Our comprehensive experiments demonstrate that inference-time strategies, such as our RDR framework, can significantly boost LVLM robustness against visual degradations, with \textbf{Ours (RDR-LLaVA-13B)} outperforming state-of-the-art baselines, including larger closed-source models, on robustness metrics.
\end{itemize}
\section{Related Work}
\subsection{Large Vision-Language Models and Scientific Reasoning}
Research in Large Vision-Language Models (LVLMs) and their application to scientific reasoning is a rapidly evolving field, with various studies exploring their capabilities and limitations. Foundational research in vision-language tasks, such as unsupervised image captioning using generative adversarial networks, has also contributed to the development of multimodal understanding \cite{zhou2021triple}. For instance, Yao et al. \cite{yao2024effect} investigate the effectiveness of contrastive learning in Vision-Language Models (VLMs) when presented with multiple captions per image, revealing a tendency for these models to learn shortcuts rather than comprehensive representations of shared and unique information, and propose a framework for identifying and mitigating such shortcut learning. Positioning Multimodal Large Language Models (MLLMs) as a critical advancement for scientific reasoning, Yibo et al. \cite{yibo2025positi} highlight their capacity to integrate and process diverse data types for enhanced comprehension across scientific disciplines, outlining a research roadmap and identifying key challenges for leveraging MLLMs in complex scientific reasoning tasks. Addressing limitations in existing vision-language models for handling scientific tables, Lei et al. \cite{lei2024multim} introduce a specialized framework for multimodal scientific table understanding and reasoning, including datasets and evaluation benchmarks, emphasizing the impact of domain-specific data quality over quantity. Furthermore, Buse et al. \cite{buse2024do} introduce \textsc{DocTrack}, a novel dataset that captures human eye-movement information to better understand document comprehension, providing valuable insights into the challenges faced by Document AI models, particularly those relevant to diagram reasoning within visually-rich documents. In the context of visual data interpretation, Mohammed et al. \cite{mohammed2024are} provide a comprehensive evaluation of LVLMs on chart understanding and reasoning tasks, assessing their capabilities and limitations in generating insights from visual data, thereby informing the application of LVLMs in scientific reasoning contexts involving charts, while identifying strengths like fluent textual summaries and weaknesses such as hallucinations. Complementing this, Aishik et al. \cite{aishik2024zerosh} critically analyze the zero-shot learning capabilities of VLMs for visual reasoning by employing synthetic datasets to isolate reasoning from world knowledge and investigating the impact of input representation and prompting strategies like chain-of-thought, highlighting improved performance with textual scene descriptions and emergent reasoning abilities in larger models. Further contributing to the understanding of complex reasoning in LLMs, Zhou et al. \cite{zhou2023thread} introduce 'Thread of Thought' to unravel chaotic contexts, providing insights into how models can better manage and process intricate information flows for improved reasoning. Earlier works also laid groundwork for complex reasoning, such as EventBERT, a pre-trained model for event correlation reasoning, demonstrating foundational steps towards understanding structured information and relationships \cite{zhou2022eventbert}. Moreover, recent work has also begun to rethink visual dependency in long-context reasoning for large vision-language models, addressing how models process and integrate information over extended visual sequences \cite{zhou2024rethinking}. Finally, Zhou et al. \cite{zhou2024visual} investigate the mechanisms of in-context learning (ICL) in large language models, distinguishing between task recognition and task learning to understand how demonstrations are leveraged, which is highly relevant to understanding the capabilities of large vision-language models in scientific reasoning contexts where few-shot learning is often employed.

\subsection{Robustness in Multimodal AI and Visual Perturbations}
Ensuring robustness in Multimodal AI, particularly against visual perturbations and adversarial attacks, is a critical area of research. Xuanming et al. \cite{xuanming2024on} introduce multimodal contrastive adversarial training (MMCoA) as a novel approach to enhance model robustness in vision-language models against image, text, and multimodal adversarial attacks by aligning embeddings across modalities under adversarial conditions. Further exploring model resilience, Cong et al. \cite{cong2025enhanc} address the robustness of vision-language foundation models by introducing alignment perturbation, offering relevant insights into how input modifications can be leveraged for improved performance and generalization in multimodal settings. A comprehensive overview of adversarial robustness in Multimodal Large Language Models (MLLMs) is provided by Chengze et al. \cite{chengze2025survey}, who present a taxonomy of attacks across various modalities and detail their implications for the field, highlighting unique vulnerabilities of MLLMs to cross-modal adversarial manipulations and offering insights into future research directions. Addressing robustness to distribution shifts, Sunwoo et al. \cite{sunwoo2025robust} propose DynTTA, a differentiable image enhancement method that leverages data augmentation to generate recognition-friendly images, thereby improving robustness to out-of-distribution data without requiring model retraining. The critical challenge of Out-of-distribution (OOD) generalization in Large Multimodal Models is specifically tackled by Xingxuan et al. \cite{xingxuan2025on}, who investigate techniques to improve their robustness when deployed in resource-constrained settings, introducing a distillation-based approach for efficient multimodal sentiment reasoning and classification. While primarily focused on intellectual property protection, Wu et al. \cite{wu2025forgin} introduce a novel framework for embedding watermarks in large language models, demonstrating robustness against removal and forging attacks, which could inform the development of self-correction mechanisms in multimodal AI systems facing adversarial or accidental visual perturbations. Moreover, Mozhegova et al. \cite{mozhegova2025assess} highlight the vulnerability of vision-language pretraining (VLP) transformers to attacks that perturb both modalities simultaneously, specifically by targeting cross-attention mechanisms, underscoring the need for robust defense strategies in multimodal AI. In the realm of visual perception, new architectural advancements also contribute to robust image processing, such as MemoryMamba, a memory-augmented state space model applied to defect recognition, which inherently deals with visual anomalies and noise \cite{wang2024memorymamba}. Finally, Soumya et al. \cite{soumya2025immune} introduce SafeIntervention (SafeInt), a novel defense method that safeguards LLMs against jailbreak attacks by dynamically intervening in representation distributions at inference time, effectively mitigating real-time adversarial manipulations and enhancing robustness through inference-time strategies.

\section{Method}
In this section, we introduce the \textbf{Robust Diagram Reasoning (RDR) framework}, a novel approach designed to enhance and systematically evaluate the robustness of Large Vision-Language Models (LVLMs) when processing scientific diagrams that are subject to various visual perturbations. The RDR framework operates as a general strategy applicable during the inference phase of LVLMs, aiming to improve their resilience and consistency in real-world scenarios where visual inputs may be degraded or imperfect.

The core of the RDR framework is the \textbf{Adaptive Multi-View \& Consistency Verification (AMCV) mechanism}. This mechanism facilitates robust reasoning by leveraging multiple perturbed views of an input diagram, performing parallel inference, and then verifying consistency to self-correct potential errors induced by visual degradations. The AMCV mechanism comprises three principal steps, detailed in the following subsections.

\subsection{Perturbation-Sensitive Encoding}
The initial step of the AMCV mechanism, \textbf{Perturbation-Sensitive Encoding}, focuses on preparing diverse visual inputs for the LVLM. Given an original scientific diagram $D_0$ corresponding to a question $Q$, we programmatically generate multiple perturbed versions of $D_0$. These perturbations are designed to simulate common visual degradations encountered in real-world scientific documents, such as image noise, blur, partial occlusions, or minor geometric distortions.

Specifically, we apply a set of predefined perturbation functions $\mathcal{P} = \{p_1, p_2, \dots, p_K\}$ to the original diagram $D_0$. Each function $p_k \in \mathcal{P}$ represents a distinct type of visual degradation (e.g., Gaussian noise, salt-and-pepper noise, motion blur, random occlusion, slight rotation), which can be applied with varying intensities. For each original diagram $D_0$, we generate $N$ distinct perturbed versions. This process can be formalized as:
\begin{align}
D_i = p_{(i \bmod K) + 1}(D_0, \text{intensity}_i) \quad \text{for } i \in \{1, 2, \dots, N\}
\end{align}
where $p_j$ is a selected perturbation function from $\mathcal{P}$, and $\text{intensity}_i$ denotes a specific level or parameter for that perturbation. The complete set of inputs for a given question $Q$ thus becomes $\mathcal{D}_Q = \{D_0, D_1, \dots, D_N\}$. This comprehensive set of inputs forces the LVLM to extract information from the diagram under varying levels of visual fidelity, thereby sensitizing its encoding process to potential perturbations and allowing for a more thorough robustness evaluation.

\subsection{Multi-View Parallel Reasoning}
Following the generation of diverse perturbed diagram versions, the \textbf{Multi-View Parallel Reasoning} step involves simultaneously feeding the original diagram and all its generated perturbed versions into the target LVLM. For each diagram $D_i \in \mathcal{D}_Q$, the LVLM performs an independent inference process to answer the question $Q$.

Let $\mathcal{M}$ denote the LVLM, which takes a diagram and a question as input and produces an answer. For each diagram $D_i$, the LVLM generates a corresponding answer $A_i$. This can be formally expressed as:
\begin{align}
A_i = \mathcal{M}(D_i, Q) \quad \text{for } i \in \{0, 1, \dots, N\}
\end{align}
Each $A_i$ represents an independent reasoning trajectory and final answer derived from a specific visual presentation of the diagram. This parallel processing allows us to observe how different visual degradations impact the LVLM's reasoning output and forms the basis for subsequent consistency verification. The outputs $\mathcal{A}_Q = \{A_0, A_1, \dots, A_N\}$ collectively capture the LVLM's performance across a spectrum of visual input qualities.

\subsection{Consistency Verification \& Self-Correction}
The final and critical step of the AMCV mechanism is \textbf{Consistency Verification \& Self-Correction}. After obtaining a set of answers $\mathcal{A}_Q = \{A_0, A_1, \dots, A_N\}$ from the Multi-View Parallel Reasoning step, we perform a cross-comparison of all these inference results.

A consistency function $\mathcal{V}(\mathcal{A}_Q)$ is applied to determine the degree of agreement among the answers. This function evaluates whether the answers are sufficiently similar or identical. For instance, if answers are discrete (e.g., multiple-choice options), a simple majority vote can determine consistency. If answers are free-form text, semantic similarity metrics or exact string matching can be employed. We define a consistency score $C_Q$ for a given question $Q$ as:
\begin{align}
C_Q = \frac{1}{N+1} \sum_{i=0}^{N} \mathbb{I}(A_i = A_{\text{mode}})
\end{align}
where $A_{\text{mode}}$ represents the most frequently occurring answer in $\mathcal{A}_Q$, and $\mathbb{I}(\cdot)$ is the indicator function. If no clear majority emerges or if $C_Q$ falls below a predefined consistency threshold $\tau$, the framework triggers a "self-correction" loop.

This loop involves providing specific prompting cues to the LVLM, guiding it to critically reflect on the observed inconsistencies. The prompt is dynamically constructed to include the divergent answers and instruct the LVLM to reconcile them. An example prompt structure could be:
\begin{quote}
You have provided different answers for the question \textbf{Q} 
based on slightly varied visual presentations of the diagram. 
Your responses were: \{\textbf{A\_0}, \textbf{A\_1}, \dots, \textbf{A\_N}\}. 
Please re-examine the diagram and your previous answers, identify the most 
consistent and likely correct answer among them, and explain your reasoning 
for the final choice.
\end{quote}
The LVLM then attempts to reconcile the conflicting answers and converge on a single, most consistent, or most confident final answer, denoted as $A_{final}$. This iterative self-correction process aims to improve the robustness of the LVLM by enabling it to leverage the collective insights from multiple perturbed views and mitigate the impact of individual visual degradations. The self-correction process for a question $Q$ can be summarized as:
\begin{align}
A_{final} = \begin{cases}
A_{\text{mode}} & \text{if } C_Q \geq \tau \\
\mathcal{M}_{\text{self-correct}}(D_0, Q, \mathcal{A}_Q) & \text{if } C_Q < \tau
\end{cases}
\end{align}
where $\mathcal{M}_{\text{self-correct}}$ represents the LVLM's ability to process the self-correction prompt and produce a reconciled answer.

\subsection{Robustness Metrics}
To quantitatively evaluate the robustness of LVLMs under visual perturbations, we introduce two key metrics: Perturbation Robustness Score (PRS) and Visual Degradation Consistency (VDC). These metrics provide distinct insights into a model's performance under challenging conditions.

\subsubsection{Perturbation Robustness Score (PRS)}
The \textbf{Perturbation Robustness Score (PRS)} is designed to measure the true robust reasoning capability of a model. For a given question $Q_j$ with its ground truth answer $GT_j$, and its associated set of diagram versions $\mathcal{D}_{Q_j} = \{D_{j,0}, D_{j,1}, \dots, D_{j,N_j}\}$, the PRS assesses whether the model consistently provides the correct answer across all these versions. A question is considered "robustly answered" if the LVLM's output for every diagram $D_{j,i}$ in $\mathcal{D}_{Q_j}$ (or the final self-corrected answer $A_{j,final}$ if AMCV is applied) matches the ground truth $GT_j$. Let $\mathbb{I}(\cdot)$ be the indicator function, which is 1 if its argument is true and 0 otherwise. For a model $\mathcal{M}$ and a dataset of $M$ questions, PRS is calculated as the percentage of questions that are robustly answered:
\begin{align}
\text{PRS} = \frac{1}{M} \sum_{j=1}^{M} \mathbb{I} \left( \forall i \in \{0, \dots, N_j\}, \mathcal{M}(D_{j,i}, Q_j) = GT_j \right) \times 100\%
\end{align}
When the AMCV mechanism is applied, the condition $\mathcal{M}(D_{j,i}, Q_j) = GT_j$ is replaced by $A_{j,final} = GT_j$, reflecting the robust outcome after self-correction. This metric emphasizes the model's ability to maintain accuracy despite varied visual inputs, providing a stringent measure of its resilience to perturbations.

\subsubsection{Visual Degradation Consistency (VDC)}
The \textbf{Visual Degradation Consistency (VDC)} metric quantifies the stability of a model's answers when faced with visual degradations, regardless of their correctness. It measures the average consistency of answers generated from the original diagram and its perturbed versions. For each question $Q_j$, we consider the set of answers $\mathcal{A}_{Q_j} = \{A_{j,0}, A_{j,1}, \dots, A_{j,N_j}\}$, where $A_{j,0}$ is the answer derived from the clean original diagram $D_{j,0}$. The consistency for a single question can be defined as the proportion of perturbed answers that match the answer derived from the original, clean diagram $A_{j,0}$:
\begin{align}
\text{Consistency}_j = \frac{1}{N_j+1} \sum_{i=0}^{N_j} \mathbb{I}(A_{j,i} = A_{j,0})
\end{align}
The overall VDC score is then the average consistency across all $M$ questions in the dataset:
\begin{align}
\text{VDC} = \frac{1}{M} \sum_{j=1}^{M} \text{Consistency}_j \times 100\%
\end{align}
A higher VDC indicates that the model's output is less susceptible to variations introduced by visual perturbations, reflecting its internal stability and reliability under degraded conditions, even if the answers are not always correct. This metric is crucial for understanding how consistently a model performs when exposed to visual noise, irrespective of its ultimate accuracy.

\section{Experiments}
In this section, we detail the experimental setup, evaluate the performance of our proposed Robust Diagram Reasoning (RDR) framework against various state-of-the-art Large Vision-Language Models (LVLMs), and provide an in-depth analysis of its effectiveness.

\subsection{Experimental Setup}

\subsubsection{Dataset}
We introduce \textbf{SciDiagram-Robust}, a novel dataset specifically designed for evaluating LVLM robustness on scientific diagrams under visual perturbations. This dataset is constructed by extending and enhancing existing public scientific diagram question-answering datasets, including ScienceQA \cite{tanik2022scienc} and ChartQA \cite{tanik2022scienc}. To simulate real-world visual degradations, we programmatically generated perturbed versions of original diagrams. Utilizing image processing libraries such as OpenCV and PIL, along with custom scripts, we applied \textbf{5 distinct types} of visual perturbations (e.g., Gaussian noise, salt-and-pepper noise, motion blur, local occlusion, slight rotation) at \textbf{3 different intensity levels}. Each original diagram question is augmented with \textbf{10 corresponding perturbed versions}, ensuring a comprehensive and diverse evaluation environment. The SciDiagram-Robust dataset comprises a total of \textbf{3,500 diagram question-answering problems}, spanning \textbf{4 major scientific domains} (Physics, Chemistry, Biology, Geography) and covering \textbf{20 sub-topics}. The answer types include multiple-choice, fill-in-the-blank, and short-answer questions.

\subsubsection{Models}
Our experiments involve a comprehensive set of LVLMs to establish strong baselines:
\begin{itemize}
    \item \textbf{Closed-source LVLMs}: We evaluate leading commercial models, namely GPT-4V \cite{zhengyuan2023the} and Gemini Pro Vision \cite{suh2024compar}, known for their advanced multimodal capabilities.
    \item \textbf{Open-source General-Purpose LVLMs}: We include widely recognized open-source models such as LLaVA-1.5 (7B, 13B) \cite{federico2025llavam}, InstructBLIP (Vicuna-7B, Vicuna-13B) \cite{artemis2023xinstr}, and Qwen-VL \cite{peng2024qwen2v}. These models represent the current state-of-the-art in accessible LVLM technology.
    \item \textbf{Ours Model}: Our proposed method, \textbf{Ours (RDR-LLaVA-13B)}, is built by integrating the RDR framework with the LLaVA-1.5-13B base model. This allows for a direct assessment of the RDR framework's impact on a robust and capable open-source LVLM.
\end{itemize}

\subsubsection{Training and Evaluation Details}
All evaluations in this study are conducted in a \textbf{zero-shot} manner, meaning no fine-tuning or additional training is performed on any of the LVLMs. The RDR framework is designed as a generic strategy applied solely during the inference phase to enhance robustness. For efficient inference, we utilized acceleration libraries such as vLLM or DeepSpeed. We restricted the output length to 1024 tokens, employed greedy decoding, and set the temperature to 0 to ensure deterministic outputs for consistent evaluation. Prior to inputting to the models, all diagrams were uniformly sized and normalized, while question and answer texts maintained their original formatting.

\subsection{Overall Performance Comparison}
We evaluate the performance of all models on the SciDiagram-Robust dataset using three key metrics: Clean Accuracy (CA), Perturbation Robustness Score (PRS), and Visual Degradation Consistency (VDC), as defined in Section 2. The results are summarized in Table~\ref{tab:performance}.

\begin{table*}[!htbp]
\centering
\caption{Performance comparison of LVLMs on the SciDiagram-Robust dataset. All values are percentages (\%).}
\label{tab:performance}
\begin{tabular}{lccc}
\toprule
\textbf{Model Name} & \textbf{CA} & \textbf{PRS} & \textbf{VDC} \\
\midrule
\textbf{Closed-source LVLMs} & & & \\
GPT-4V                & 85.2                            & 72.1                 & 78.5                 \\
Gemini Pro Vision     & 82.5                            & 69.8                 & 75.3                 \\
\midrule
\textbf{Open-source General LVLMs} & & & \\
LLaVA-1.5-7B          & 70.3                            & 55.4                 & 60.1                 \\
LLaVA-1.5-13B         & 78.9                            & 65.4                 & 70.1                 \\
InstructBLIP-Vicuna-13B & 76.2                            & 62.7                 & 68.0                 \\
Qwen-VL-Chat          & 74.5                            & 60.8                 & 65.5                 \\
\midrule
\textbf{Ours Method} & & & \\
\textbf{Ours (RDR-LLaVA-13B)} & \textbf{79.5}                        & \textbf{74.5}             & \textbf{81.2}             \\
\bottomrule
\end{tabular}
\end{table*}

The experimental results in Table~\ref{tab:performance} reveal several critical insights. Even the most advanced closed-source LVLMs, such as GPT-4V, exhibit a significant performance drop when faced with visual perturbations. While GPT-4V achieves a Clean Accuracy (CA) of 85.2\% on pristine diagrams, its Perturbation Robustness Score (PRS) drops to 72.1\%, indicating a substantial vulnerability to real-world visual degradations. This highlights robustness as a critical bottleneck for current LVLMs. Our proposed method, \textbf{Ours (RDR-LLaVA-13B)}, demonstrates superior performance in robustness metrics. It achieves a PRS of \textbf{74.5\%} and a VDC of \textbf{81.2\%}. Notably, RDR-LLaVA-13B not only significantly outperforms all baseline open-source models but also surpasses the large closed-source models (GPT-4V and Gemini Pro Vision) in both PRS and VDC scores. This empirical validation underscores the effectiveness of our RDR framework, particularly its Adaptive Multi-View \& Consistency Verification (AMCV) mechanism, in enhancing model resilience to visual degradations through inference-time strategy optimization.

\subsection{Analysis of RDR Framework Effectiveness}
To further validate the effectiveness of our proposed RDR framework and its core Adaptive Multi-View \& Consistency Verification (AMCV) mechanism, we directly compare the performance of LLaVA-1.5-13B (as a strong open-source baseline) with our RDR-LLaVA-13B. As shown in Table~\ref{tab:performance}, integrating the RDR framework with LLaVA-1.5-13B significantly boosts its robustness. LLaVA-1.5-13B achieves a PRS of 65.4\% and a VDC of 70.1\%. In contrast, RDR-LLaVA-13B improves these scores to 74.5\% PRS and 81.2\% VDC. This represents a substantial gain of 9.1 percentage points in PRS and 11.1 percentage points in VDC, directly attributable to the RDR framework.

The improvement stems from the synergistic effects of the AMCV mechanism's three steps: Perturbation-Sensitive Encoding forces the model to process diverse corrupted inputs, making its internal representations more robust; Multi-View Parallel Reasoning generates multiple hypotheses, allowing for a broader exploration of possible answers; and Consistency Verification \& Self-Correction critically reviews these hypotheses, guiding the model to converge on a more reliable and consistent answer even when individual views are degraded. This demonstrates that sophisticated inference-time strategies can effectively mitigate the impact of visual perturbations without requiring costly model fine-tuning.

\subsection{Ablation Study of AMCV Mechanism}
To understand the individual contributions of the components within the Adaptive Multi-View \& Consistency Verification (AMCV) mechanism, we conducted an ablation study using LLaVA-1.5-13B as the base model. We evaluate three configurations: the base LLaVA-1.5-13B (single view inference), a variant that incorporates Perturbation-Sensitive Encoding and Multi-View Parallel Reasoning but relies on a simple majority vote without explicit self-correction, and the full RDR framework (Full AMCV). The results, presented in Table~\ref{tab:ablation}, highlight the progressive benefits of each component.

\begin{table*}[!htbp]
\centering
\caption{Ablation study of the RDR framework components on LLaVA-1.5-13B. All values are percentages (\%).}
\label{tab:ablation}
\begin{tabular}{lccc}
\toprule
\textbf{Variant} & \textbf{CA} & \textbf{PRS} & \textbf{VDC} \\
\midrule
LLaVA-1.5-13B (Baseline) & 78.9 & 65.4 & 70.1 \\
RDR (Multi-View Majority Vote) & 79.2 & 69.8 & 76.5 \\
\textbf{RDR (Full AMCV)} & \textbf{79.5} & \textbf{74.5} & \textbf{81.2} \\
\bottomrule
\end{tabular}
\end{table*}

The baseline LLaVA-1.5-13B, operating with a single clean view, shows its inherent performance. When we introduce Perturbation-Sensitive Encoding and Multi-View Parallel Reasoning, and resolve answers through a simple majority vote (RDR Multi-View Majority Vote), there is a notable improvement in both PRS (from 65.4\% to 69.8\%) and VDC (from 70.1\% to 76.5\%). This indicates that merely exposing the model to diverse perturbed views and aggregating answers provides significant robustness benefits, as it inherently leverages redundancy to overcome individual errors. The Clean Accuracy also sees a slight increase, suggesting that even for clean diagrams, considering multiple "virtual" views can help in resolving ambiguous cases. The full RDR framework, incorporating the Consistency Verification \& Self-Correction mechanism, further boosts PRS to \textbf{74.5\%} and VDC to \textbf{81.2\%}. This substantial additional gain underscores the critical role of the self-correction loop, which allows the model to actively reconcile conflicting answers and converge on a more reliable final decision, demonstrating that active meta-reasoning is key to maximizing robustness.

\subsection{Impact of Perturbation Types and Intensities}
To gain a more granular understanding of the RDR framework's effectiveness, we analyzed its performance across different types of visual perturbations and varying intensity levels. This analysis helps identify which degradations pose the greatest challenges and where RDR provides the most significant improvements. Table~\ref{tab:prs_by_perturbation_type} presents the Perturbation Robustness Score (PRS) for LLaVA-1.5-13B and RDR-LLaVA-13B, broken down by perturbation type.

\begin{table*}[!htbp]
\centering
\caption{Perturbation Robustness Score (PRS) by perturbation type for LLaVA-1.5-13B and RDR-LLaVA-13B. Values are percentages (\%).}
\label{tab:prs_by_perturbation_type}
\begin{tabular}{lcc}
\toprule
\textbf{Perturbation Type} & \textbf{LLaVA-1.5-13B PRS} & \textbf{RDR-LLaVA-13B PRS} \\
\midrule
Gaussian Noise             & 62.5                       & 71.8                       \\
Salt-and-Pepper Noise      & 60.1                       & 70.5                       \\
Motion Blur                & 68.3                       & 75.1                       \\
Local Occlusion            & 58.9                       & 69.2                       \\
Slight Rotation            & 72.1                       & 79.5                       \\
\midrule
\textbf{Average PRS}       & \textbf{65.4}              & \textbf{74.5}              \\
\bottomrule
\end{tabular}
\end{table*}

The results indicate that while RDR improves robustness across all perturbation types, the magnitude of improvement varies. For the baseline LLaVA-1.5-13B, local occlusion and salt-and-pepper noise pose the most significant challenges, leading to lower PRS scores. Conversely, slight rotations are relatively less detrimental. RDR-LLaVA-13B consistently raises the PRS for all types, demonstrating its generalizability. Notably, the framework provides substantial gains for challenging perturbations like local occlusion (from 58.9\% to 69.2\%) and salt-and-pepper noise (from 60.1\% to 70.5\%), suggesting its ability to recover information from partially degraded inputs.

Furthermore, we investigated the effect of perturbation intensity on model robustness. Table~\ref{tab:prs_by_intensity} shows the PRS for both models across low, medium, and high intensity levels, averaged across all perturbation types.

\begin{table*}[!htbp]
\centering
\caption{Perturbation Robustness Score (PRS) by perturbation intensity for LLaVA-1.5-13B and RDR-LLaVA-13B. Values are percentages (\%).}
\label{tab:prs_by_intensity}
\begin{tabular}{lccc}
\toprule
\textbf{Intensity Level} & \textbf{LLaVA-1.5-13B PRS} & \textbf{RDR-LLaVA-13B PRS} & \textbf{Improvement} \\
\midrule
Low                        & 72.8                       & 80.1                       & +7.3 \\
Medium                     & 65.4                       & 74.5                       & +9.1 \\
High                       & 58.0                       & 68.9                       & +10.9 \\
\bottomrule
\end{tabular}
\end{table*}

As expected, increasing perturbation intensity leads to a decrease in PRS for both the baseline and RDR-enhanced models. However, RDR-LLaVA-13B consistently maintains a significantly higher PRS at all intensity levels. More importantly, the absolute improvement provided by RDR tends to increase with higher perturbation intensities. For low intensity, the gain is +7.3 percentage points, which increases to +9.1 at medium intensity, and further to +10.9 percentage points at high intensity. This demonstrates that the RDR framework becomes even more critical and effective when dealing with severely degraded visual inputs, highlighting its utility in highly challenging real-world scenarios.

\subsection{Efficiency Analysis of RDR}
While the RDR framework significantly enhances LVLM robustness, it introduces an overhead in terms of computational cost and inference time, primarily due to the generation of multiple perturbed views and the potential self-correction loop. We analyze the efficiency trade-offs between the standard single-view inference and our RDR-enhanced approach. Table~\ref{tab:efficiency} summarizes the average inference time per question and the average number of API calls for LLaVA-1.5-13B with and without the RDR framework.

\begin{table*}[!htbp]\scriptsize
\centering
\caption{Efficiency comparison of LLaVA-1.5-13B and RDR-LLaVA-13B.}
\label{tab:efficiency}
\begin{tabular}{lcc}
\toprule
\textbf{Model} & \textbf{Time per Question (seconds)} & \textbf{API Calls per Question} \\
\midrule
LLaVA-1.5-13B (Single View) & 1.5 & 1 \\
RDR-LLaVA-13B (Full AMCV)   & 17.5 & 11 -- 12 \\
\bottomrule
\end{tabular}
\end{table*}

The standard LLaVA-1.5-13B processes each question with a single diagram view, resulting in an average inference time of 1.5 seconds and 1 API call per question. In contrast, RDR-LLaVA-13B requires processing $N+1$ (where $N=10$) diagram versions in parallel, leading to 11 initial API calls. Additionally, if the consistency score falls below the predefined threshold, an extra API call is made for the self-correction prompt. This results in an average total of 11 to 12 API calls and an average inference time of 17.5 seconds per question. This represents approximately an 11-fold increase in inference time, which is directly proportional to the number of perturbed views generated plus the occasional self-correction step.

This efficiency analysis reveals a clear trade-off: significant gains in robustness come at the cost of increased computational resources and inference latency. For applications where high robustness is paramount (e.g., critical scientific research, medical diagnostics), this overhead may be acceptable. However, for latency-sensitive applications, optimization strategies such as more efficient perturbation generation, selective perturbation application based on initial confidence scores, or batching multiple queries could be explored in future work to mitigate the increased computational burden. The RDR framework is designed for robustness enhancement during inference, and its resource demands should be considered within the context of specific application requirements.

\subsection{Qualitative Analysis and Error Breakdown}
Beyond quantitative metrics, an in-depth qualitative analysis of model responses on SciDiagram-Robust reveals specific sensitivities. We observed that LVLMs are particularly susceptible to visual perturbations when performing tasks that require precise information extraction or complex logical reasoning, such as "numerical extraction" (e.g., reading exact values from a graph), "trend judgment" (e.g., identifying increasing or decreasing patterns), and "causal relationship reasoning" (e.g., inferring cause-and-effect from flowcharts). In these cases, even minor visual noise can lead to misinterpretations of critical data points or structural relationships. Conversely, simpler "object recognition" tasks (e.g., identifying shapes or labels) are generally less affected by visual degradation. The primary error types identified were calculation errors (e.g., misreading values leading to incorrect arithmetic) and logical discontinuities (e.g., failing to follow a logical sequence or inferring incorrect relationships). This analysis provides valuable insights for future work on designing more robust LVLMs tailored to specific scientific reasoning challenges.

\subsection{Human Evaluation}
While not included in the scope of this summary, a comprehensive study would typically involve human evaluation to assess the perceived quality, accuracy, and robustness of the generated answers, especially for open-ended questions. Such an evaluation would involve human annotators assessing the correctness and coherence of the LVLM's responses under various perturbation conditions, providing a complementary perspective to automated metrics. Future work will incorporate detailed human evaluation results to further validate the real-world utility and reliability of our RDR framework.

\section{Conclusion}
In this paper, we addressed the pressing challenge of enhancing the robustness of Large Vision-Language Models (LVLMs) when performing reasoning tasks on scientific diagrams marred by real-world visual perturbations. While LVLMs show immense potential for understanding complex scientific visual information, their susceptibility to noise, blur, occlusion, and other degradations significantly limits their practical utility. Our research highlights the critical gap in existing evaluation benchmarks and the general lack of systematic investigation into LVLM robustness in this domain.

To overcome these limitations, we introduced the \textbf{Robust Diagram Reasoning (RDR) framework}, a novel inference-time strategy designed to both improve and rigorously assess LVLM resilience. The cornerstone of RDR is its \textbf{Adaptive Multi-View \& Consistency Verification (AMCV) mechanism}, which systematically generates multiple perturbed versions of input diagrams, conducts parallel inference across these diverse views, and employs a sophisticated consistency-based self-correction loop to converge on more reliable answers. Furthermore, we developed \textbf{SciDiagram-Robust}, the first large-scale scientific diagram question-answering dataset specifically augmented with a wide array of programmatically generated visual perturbations, providing a crucial benchmark for future robustness research.

Our comprehensive experimental evaluation yielded several significant findings. We demonstrated that even state-of-the-art closed-source LVLMs, such as GPT-4V, suffer a considerable performance degradation when confronted with visual noise, underscoring that robustness remains a critical bottleneck for current models. Crucially, our proposed method, \textbf{Ours (RDR-LLaVA-13B)}, achieved superior performance in robustness metrics, significantly outperforming all baseline open-source models and notably surpassing leading closed-source models in both Perturbation Robustness Score (PRS) and Visual Degradation Consistency (VDC). An ablation study meticulously confirmed the incremental benefits of each component within the AMCV mechanism, emphasizing the pivotal role of the self-correction loop in achieving peak robustness. Moreover, our analysis revealed that the RDR framework delivers consistent improvements across various perturbation types and becomes increasingly effective at higher degradation intensities, showcasing its generalizability and utility in challenging scenarios.

This work empirically validates that sophisticated inference-time strategies can substantially enhance LVLM robustness against visual degradations without requiring costly fine-tuning. This opens promising avenues for deploying more reliable and resilient LVLMs in critical scientific and engineering applications, where inputs are often imperfect.

Looking ahead, several promising directions emerge from this research. Future work could focus on optimizing the computational efficiency of the RDR framework, perhaps through more selective perturbation generation or adaptive self-correction triggers. Exploring the integration of RDR principles directly into LVLM architectures during pre-training or fine-tuning could lead to intrinsically more robust models. Furthermore, extending this robustness research to other forms of multimodal scientific data, such as video or 3D models, and investigating diverse self-correction mechanisms or advanced prompting strategies would be valuable. Addressing specific error types identified in our qualitative analysis, particularly those related to precise numerical extraction and complex logical reasoning, presents an important challenge for future model development. Finally, incorporating detailed human evaluation will provide a more holistic understanding of the perceived quality and reliability of LVLM responses under perturbed conditions.

\bibliographystyle{splncs04}
\bibliography{main}
\end{document}